\documentclass[journal]{IEEEtran}
\usepackage[compatibility=false]{caption}
\usepackage{xcolor,soul,framed} 

\colorlet{shadecolor}{yellow}
\usepackage[pdftex]{graphicx}
\graphicspath{{../pdf/}{../jpeg/}}
\DeclareUnicodeCharacter{202F}{\,} 
\DeclareGraphicsExtensions{.pdf,.jpeg,.png}

\usepackage[cmex10]{amsmath}
\usepackage{array}
\usepackage{mdwmath}
\usepackage{mdwtab}
\usepackage{eqparbox}
\usepackage{url}
\usepackage{hyperref}
\usepackage{tikz}
\usetikzlibrary{arrows.meta, positioning, shapes.geometric}
\usepackage[justification=centering,compatibility=false]{caption}
\usepackage[cmex10]{amsmath}
\usepackage{amssymb}
\usepackage{algorithm}
\usepackage{algorithmic}
\usepackage[percent]{overpic}
\usepackage{placeins}
\usepackage{float}
\usepackage{stfloats}  
\usepackage{afterpage}
\usepackage{mdwmath}
\usepackage{multirow}
\usepackage{float} 

\begin{document}
\title{Human Activity Recognition via Ultra-Wideband Data: 
A Framework for Dimensionality Reduction, Pattern Discovery, 
and Predictive Modeling}

\author{
Nahid~Sahel~Gozin,
Reza~Sedaghat,~\IEEEmembership{Senior~Member,~IEEE,}
Prathap~Siddavaatam,~\IEEEmembership{Member,~IEEE,}

\thanks{All authors are with the OPR-AL Lab, Toronto Metropolitan University, 350 Victoria Street, Toronto, ON M5B 2K3, Canada.}

\thanks{E-mail: nsahelgozin@ee.torontomu.ca;
rsedagha@ee.torontomu.ca;  prathap.siddavaatam@ee.torontomu.ca}%
\thanks{Project website: https://www.ee.torontomu.ca/opr/}
}


\maketitle
\begin{abstract}   
Recent advances in sensor technology have enabled more effective human activity recognition (HAR), particularly in real time systems with limited computational resources. However, Ultra-Wideband (UWB) radar data remain challenging due to high dimensionality, noise, complexity, and non-linear characteristics. This research proposes a framework to efficiently reduce data size, uncovers significant patterns, and classifies six activity types (Standff, Liedown, Noactivity, Sit, Stand, and Walk) from UWB signals with high accuracy. Two novel dimensionality reduction techniques are introduced in this paper. The first, Clustered Polynomial Expansion with Incremental PCA (CPE-IPCA), combines clustering and polynomial feature expansion with Incremental PCA, preserving 100\% of the variance in only 50 components. The second, Post-PCA Standardization Approach (PPSA), standardizes data after PCA and retains 99.1\% of the variance in 80 components, achieving superior compression and computational efficiency compared to conventional non-linear methods. Frequent patterns are identified using Apriori and FP-Growth, which are then classified with Random Forest and a Vector Space Model (VSM). The framework achieves 100\% accuracy with Random Forest on CPE-IPCA and 99\% on PPSA, while VSM attains 100\% precision, recall, and F1 on PPSA and near-perfect performance on CPE-IPCA (precision 1.00, recall 0.98–1.00, F1 0.99–1.00), demonstrating a fast, interpretable, and robust HAR system suitable for healthcare, assisted living, and smart environments.
\end{abstract}

\section{Introduction}
Human Activity Recognition, or HAR, has become an increasingly important field because it can be applied in many areas, from healthcare and security to entertainment, the military, and smart environments~\cite{Daga2022, Hussain2020, Yousaf2024, khalid2022multi}. One of the main technologies used for HAR is Ultra-Wideband (UWB) radar~\cite{Khan2025, wang2022gcn, Yousaf2024, Maitre2021, Yin2024, noori2021uwb}. Unlike wearable sensors, UWB does not require people to carry or wear any device, making it more convenient and less intrusive.

UWB has several benefits. It helps preserve privacy, works well even in dark or obstructed areas~\cite{li2024advancing, Yousaf2024, wang2022gcn}, and can track movement over time. At the same time, UWB data can be difficult to handle because it is often high-dimensional, noisy, and non-linear ~\cite{Zhu2022_continues, lafontaine2023denoising, Maitre2023}.

Traditional techniques such as Principal Component Analysis (PCA)~\cite{Yin2024, noori2021uwb}, Convolutional Autoencoders (CAEs)~\cite{lafontaine2023denoising}, and t-SNE~\cite{khan2024} can help reduce the size of the data. Still, the features they produce are sometimes hard to interpret. They may also be too slow or resource-heavy for real-time applications or devices with limited computing power ~\cite{khan2024}.

To address these challenges, we introduce two new novel reduction methods. The first is Clustered Polynomial Expansion with Incremental PCA (CPE-IPCA), which groups related features, generates polynomial terms, and applies Incremental PCA. This method keeps the most important information while also capturing nonlinear patterns. The second method, Post-PCA Standardization (PPSA), applies a simple normalization step after PCA, making the results easier to understand without losing key details.

We also use Apriori and FP-Growth ~\cite{semerci2022thoughts,
mudumba2024mine} to identify rules connecting principal components to human activities. These rules are combined with Random Forest ~\cite{thakur2023grrf} and Vector Space Models ~\cite{sunori2024knn} to classify activities. By linking dimensionality reduction, rule discovery, and classification, our approach achieves both high accuracy and clear interpretability.

The main contributions of this work are three key aspects. First, we introduce CPE-IPCA, which captures complex, nonlinear patterns in UWB data. Second, {PPSA} improves PCA results by making them more interpretable. Finally, we show that combining association rule mining with classifiers not only increases accuracy in activity recognition but also makes it easier to understand the reasoning behind the decisions.

\section{Related Work}

One widely used device-free method is Ultra-Wideband (UWB) radar~\cite{Sharma2019, schmidhammer2022, GhosianMoghaddam2025, Yakoub2024, Li2024}. UWB can detect movement without touching the person~\cite{Sharma2019}, is simple to install~\cite{schmidhammer2022}, and can sense motion through walls~\cite{schmidhammer2022, Yakoub2024}. This has been described as less intrusive in terms of privacy \cite{GhosianMoghaddam2025}. However, UWB has some drawbacks. Very small or accidental movements can trigger false alarms, and noise or signals from other devices can reduce accuracy~\cite{Li2024}, and it is also described as less intrusive in terms of privacy \cite{GhosianMoghaddam2025}.

Most research using UWB has been done in the labs and usually focuses on single actions. Real world situations are harder because activities take different amounts of time, may overlap, and need to be recognized quickly using several sensors~\cite{Ullmann2023}.

To improve the challenges, researchers have tried several methods. Dimensionality reduction makes data easier to handle while keeping the important information~\cite{khan2024, rani2024comparative}. Machine learning and deep learning have been applied to increase accuracy~\cite{WangBasiri2024, Yakoub2024}. Fuzzy systems have been used to manage uncertainty and situations with more than one person~\cite{PoloRodriguez2023}.

Deep learning has given strong results. Yakoub et al.~\cite{Yakoub2024} used a two-dimensional CNN and reached 99.6\% accuracy. Polo et al.~\cite{Polo2025} combined UWB fingerprinting with CNN and LSTM to track daily activities in homes with high precision. Li et al.~\cite{Li2024} used an Attention LSTM with CNN to detect movement and vital signs like breathing and heart rate, reaching over 90\% accuracy for fall detection.

Other researchers have also made improvements. Zhao et al.~\cite{Zhao2022} showed that using data from several UWB radars improves results: ResNet reached 84.1\% and AlexNet 80.3\%. Khan et al.~\cite{Khan2025} developed a hybrid model, HDL4AR, that combines LSTM and 1D-CNN. With 22 people performing ten activities, the model reached nearly 98.5\% accuracy. Noori et al.~\cite{noori2021uwb} applied LSTM with PCA and LDA to simplify data and reported 99.6\% accuracy. Maitre et al.~\cite{Maitre2023} tested three UWB radars in a smart apartment and found Chebyshev filtering improved accuracy by 30\%. Abudalfa and Bouchard~\cite{Abudalfa2025} designed a hybrid ResNet-18/XCiT model for fall detection, reaching 99\% accuracy but requiring strong computing power.

Some studies also focused on UWB positioning. Kim and Pyun~\cite{KimPyun2024} introduced a denoising two-way ranging method that reduced errors by 34\%. Lu et al.~\cite{LuSheuKuo2021} combined LSTM and DNN models with a weighted positioning approach, improving accuracy for both line-of-sight and non-line-of-sight situations.

These studies show that UWB radar can be useful for human activity recognition. At the same time, they highlight that recognizing activities reliably in real-world situations with multiple sensors and continuous monitoring is still difficult.

Another research approach has concentrated on finding patterns directly. Association Rule Mining (ARM) has been applied to large, high-dimensional datasets to uncover meaningful relationships~\cite{He2023, Zhiyuan2024}. It has been useful in many fields~\cite{Zhang2025, T2DM_ARM2024, Zou2025}, and Li and Wang~\cite{LiWang2024} pointed out its strength in large-scale applications. Newer versions such as DT-CARMA have shown that rule-based models can even improve classification accuracy~\cite{Lu2024}, which makes ARM a valuable tool for interpretable HAR pipelines.

Classification remains at the heart of HAR research. Ullmann et al.~\cite{Ullmannclass2023} used a multi-label ResNet-50 with fused radar data to classify multiple human activities, achieving 95.8\% accuracy and a 92.08\% F1 score. However, the method is limited by fixed-length windowing, high computational cost, overlapping motions, sensitivity to the classification threshold, and the lack of runtime performance reporting.

Feng et al. (2023) found that using Random Forest helps recognize human activities more accurately. By building many decision trees together, the method can work with complex data and avoids overfitting. It achieved 98\% accuracy on the test set and 95.5\% with cross-validation. However, it requires more training time, can be difficult to interpret, and the model can become large \cite{Feng2023}.

Ezzeldin et al.\cite{Ezzeldin2024} proposed a hybrid hierarchical approach combining Transformers with traditional ML (RF, SVM, KNN, LR) for multi-modal HAR, achieving 99.69–97.4\% accuracy on PAMAP2, CASAS, UCI HAR, and UCI HAPT, but it is quite complex, memory-intensive, and requires significant computing power.

Several studies have compared classical classifiers (KNN, SVM, Logistic Regression, RF) with deep networks (LSTM, Transformers) for Human Activity Recognition. Transformers often achieve the highest accuracy but are computationally demanding. Random Forest methods applied to UWB signals are robust to noise and high-dimensional data, but they can be slower to train, harder to interpret, and less suitable for real-time use \cite{Halim2022}.

These studies show that it is still hard to extract complex features, understand how the models make decisions, and use them in real-time applications for UWB-based human activity recognition.

\section{System Model and Proposed Framework}

Human Activity Recognition (HAR) relies on extracting meaningful information from high-dimensional sensor data. This research presents a robust HAR pipeline using Ultra-Wideband (UWB) signals, built on two independent, novel dimensionality reduction methods that capture non-linear feature relationships. CPE-IPCA models non-linear dependencies by clustering correlated features, applying polynomial expansion, and performing scalable incremental PCA. PPSA-NLD, applied separately, standardizes PCA-projected components and incorporates a normalized discriminant stage, enhancing interpretability and discriminative power beyond conventional PCA. These reduced features are subsequently processed through association rule mining and classified using Random Forest (RF) and Vector Space Model (VSM) classifiers to recognize six distinct human activities with high accuracy.

\subsection{UWB Data Acquisition and Preprocessing}
For this research, we utilized the Ultra Wideband (UWB) dataset~\cite{uwb_dataset} to capture high resolution spatiotemporal information for real time Human Activity Recognition (HAR). Data were collected in controlled indoor environments with multiple participants performing diverse movements, enabling the capture of rich motion patterns. Measurements were obtained from several synchronized anchor nodes, providing spatial diversity that enhances activity modeling accuracy~\cite{uwb_dataset}.

The raw Channel Impulse Response (CIR) data were provided as complex-valued matrices. During preprocessing, missing and duplicate entries were removed, and the complex numbers were separated into real and imaginary components for efficient numerical handling. Each sample was annotated with one of six activity classes: \textit{noactivity}, \textit{liedown}, \textit{standff}, \textit{sit}, \textit{stand}, or \textit{walk}. Table~\ref{tab:dataset} summarizes the sample distribution, with “No Activity” being the most frequent and “Standff” the least.

\begin{table}[htbp]
\caption{Activity Class Distribution}
\centering
\begin{tabular}{lc}
\hline
Activity Class & Number of Instances \\
\hline
No Activity & 7,032 \\
Lie Down & 1,729 \\
Standff & 1,668 \\
Sit & 2,230 \\
Stand & 2,236 \\
Walk & 1,898 \\
\hline
Total & 17,063 \\
\hline
\end{tabular}
\label{tab:dataset}
\end{table}

\subsection{Dimensionality Reduction Techniques}

UWB data~\cite{uwb_dataset} poses significant challenges for Human Activity Recognition (HAR) due to its high dimensionality and non-linear nature. To address these issues, we propose two novel dimensionality reduction methods: CPE-IPCA and PPSA. CPE-IPCA efficiently captures non-linear dependencies while preserving structural information, whereas PPSA enhances PCA through post-projection standardization, improving stability and interpretability. Both methods yield compact, information-rich representations that enhance computational efficiency and classification accuracy.

\subsubsection{Clustered Polynomial Expansion with Incremental PCA}

Our proposed CPE-IPCA is a novel hierarchical non-linear dimensionality reduction framework for high-dimensional UWB data. It integrates feature clustering, polynomial expansion, incremental PCA, and post-IPCA standardization to yield compact, informative features with reduced computational cost. The full procedure is detailed in Algorithm~\ref{alg:cpe-ipca1}.

\begin{algorithm}[h]
\caption{CPE-IPCA Feature Reduction Algorithm}
\label{alg:cpe-ipca1}
\begin{algorithmic}[1]
\REQUIRE Feature matrix $X \in \mathbb{R}^{m \times n}$, number of clusters $K$, polynomial expansion degree $p$, number of PCA components $r$, variance threshold $\theta$
\ENSURE Reduced feature matrix $Z \in \mathbb{R}^{m \times r}$
\STATE Standardize the input features to zero mean and unit variance
\STATE Compute the absolute correlation matrix of the features
\STATE Cluster features into $K$ groups based on correlation similarity
\FOR{each cluster $k = 1, \dots, K$}
    \STATE Compute the cluster representative by averaging member features
    \STATE Apply polynomial expansion of degree $p$ to the cluster features
\ENDFOR
\STATE Generate cross-cluster interaction features
\STATE Concatenate all expanded features into a single matrix of size $m \times d$
\STATE Apply Incremental PCA to obtain a reduced-dimensional representation of size $m \times r$
\STATE Standardize the principal components
\STATE Check cumulative explained variance; adjust $r$ or $p$ if necessary to satisfy $\theta$
\RETURN Reduced feature matrix $Z$
\end{algorithmic}
\end{algorithm}

Our proposed CPE-IPCA framework compresses high dimensional UWB data via a structured transformation that preserves non-linear dependencies while maintaining interpretability and scalability.

\noindent\textbf{a) Feature Clustering:}  
Initially, the correlation structure among features is captured using the absolute correlation matrix as follows:
\begin{equation}
\begin{aligned}
R &= \lvert \mathrm{corr}(X) \rvert = [r_{ij}]_{n \times n}, \\
r_{ij} &= \frac{\mathrm{Cov}(X_i, X_j)}{\sigma_{X_i}\sigma_{X_j}}.
\end{aligned}
\label{eq:corr_matrix}
\end{equation}

\noindent
Equation~\eqref{eq:corr_matrix} defines the computation of the correlation matrix, where 
$X \in \mathbb{R}^{m \times n}$ is the dataset, 
$r_{ij}$ is the correlation between features $X_i$ and $X_j$, 
$\mathrm{Cov}(X_i, X_j)$ is the covariance, 
$\sigma_{X_i}$ and $\sigma_{X_j}$ are the standard deviations, and 
$R$ is the absolute correlation matrix.

To minimize redundancy, features exhibiting similar correlation profiles are grouped into K clusters using the k-means clustering algorithm. 

The membership of features in each cluster is represented by $C_k$ and defined in Equation~\eqref{eq:cluster_def}:
\begin{equation}
\label{eq:cluster_def}
C_k = \{ X_j \mid k = \arg\min_{k} \| R_j - \mu_{k} \| \}, \quad k = 1, 2, \dots, K
\end{equation}
where (Eq.~\ref{eq:cluster_def}):  
$X_j \in \mathbb{R}^m$ denotes the $j^{\text{th}}$ feature vector,  
$R_j \in \mathbb{R}^n$ is the correlation vector of $X_j$ with all other features,  
$\mu_k \in \mathbb{R}^n$ represents the centroid of cluster $C_k$ in the correlation space,  
$k$ is the cluster index, and  
$K$ is the total number of clusters.  

Once the feature clusters $\{ C_k \}_{k=1}^{K}$ are formed, a representative feature is computed for each cluster across all samples as defined in Equation~\eqref{eq:cluster_rep}:
\begin{equation}
\label{eq:cluster_rep}
z_k(i) = \frac{1}{|C_k|} \sum_{j \in C_k} x_{ij}, \quad i = 1, 2, \dots, m
\end{equation}
where (Eq.~\ref{eq:cluster_rep}):  
$z_k(i)$ denotes the representative value of cluster $C_k$ for sample $i$,  
$|C_k|$ is the number of features in cluster $C_k$, and  
$x_{ij}$ is the value of feature $j$ in sample $i$.  

Thus, Equation~\eqref{eq:cluster_rep} employs the clustered feature sets $C_k$ derived from Equation~\eqref{eq:cluster_def} to compute representative features, effectively reducing the dataset’s dimensionality while maintaining correlation-based structure.

The cluster-wise reduced matrix is then formed as:
\begin{equation}
\label{eq:cluster_matrix}
X_C = [z_1, z_2, \ldots, z_K] \in \mathbb{R}^{m \times K}
\end{equation}

\noindent\textit{where (Eq.~\ref{eq:cluster_matrix})}  
$X_C$ is the cluster representative matrix containing $K$ intermediate features for all $m$ samples.  
Here, $K$ (Eqs.~\ref{eq:cluster_def} and~\ref{eq:cluster_matrix}) denotes the total number of feature clusters, which determines the intermediate representation before polynomial expansion and Incremental PCA.


The proposed CPE-IPCA framework employs correlation-based clustering for early-stage compression, with novelty in the integrated steps of feature clustering (Eq.~\eqref{eq:cluster_def}), computing cluster representatives (Eq.~\eqref{eq:cluster_rep}), and forming the reduced matrix (Eq.~\eqref{eq:cluster_matrix}), preserving inter-feature relationships while reducing dimensionality for more effective non-linear modeling.



\noindent\textbf{b) Polynomial Expansion:}
To capture non-linear dependencies among clusters, the cluster level representative features are expanded using a polynomial mapping:

\begin{equation}
\Phi(\mathbf{z}_i) = [z_{i1}, z_{i2}, \ldots, z_{iK}, z_{i1}z_{i2}, z_{i1}z_{i3}, \ldots, z_{i(K-1)}z_{iK}]
\label{eq:poly_expansion}
\end{equation}

In Eq.~\eqref{eq:poly_expansion}, $\mathbf{z}_i = [z_{i1}, z_{i2}, \ldots, z_{iK}] \in \mathbb{R}^{K}$ represents the $K$-dimensional feature vector for the $i^{\text{th}}$ sample, where $z_{ij}$ denotes the representative value of the $j^{\text{th}}$ cluster for sample $i$ $(j = 1, \ldots, K)$. The mapping $\Phi(\mathbf{z}_i)$ produces a vector that includes both the original (linear) cluster features $z_{ij}$ and all pairwise product terms $z_{ij}z_{ik}$ for $1 \le j < k \le K$, thereby capturing non-linear dependencies among the cluster representative features.

We define the polynomially expanded feature matrix as:
\begin{align}
X_P &= \Phi(X_C) \in \mathbb{R}^{m \times d}, \nonumber \\
d   &= K + \frac{K(K-1)}{2}
\label{eq:poly_matrix}
\end{align}

In Eq.~\eqref{eq:poly_matrix}, $X_P$ is the polynomially expanded feature matrix with $m$ samples and $d$ features derived from the $K$ cluster representatives in Eq.~\eqref{eq:cluster_def}.

The key novelty of this step is applying polynomial expansion only to the cluster-representative features ($\mathbf{z}_i$ in Eq.~\eqref{eq:poly_expansion}), preserving nonlinear inter-cluster relationships while reducing dimensional growth and computational cost compared to conventional PCA and its nonlinear variants.

\noindent\textbf{c) Incremental PCA (IPCA):}  
Batch-wise mean and covariance updates are performed as:
\begin{equation}
\mu_b = \frac{N_{b-1}\mu_{b-1} + m_b \bar{x}_b}{N_{b-1} + m_b}
\label{eq:ipca_mean}
\end{equation}
\noindent\textit{where (Eq.~\ref{eq:ipca_mean})}  
$\mu_b$ is the updated batch mean,  
$\mu_{b-1}$ is the previous mean,  
$N_{b-1}$ is the cumulative number of samples processed,  
$m_b$ is the batch size,  
and $\bar{x}_b$ is the mean of the current batch.  

The batch-wise covariance update is computed as:
\begin{equation}
S_b = \frac{N_{b-1}}{N_{b-1} + m_b} S_{b-1} + \frac{N_{b-1}m_b}{(N_{b-1} + m_b)^2} C_b
\label{eq:ipca_cov}
\end{equation}
\noindent\textit{where (Eq.~\ref{eq:ipca_cov})}  
$S_b$ is the updated covariance matrix,  
$S_{b-1}$ is the covariance matrix from the previous batch,  
and $C_b$ is the batch covariance adjustment term defined in Eq.~\eqref{eq:ipca_cb}.  

The batch adjustment term $C_b$ is calculated as:
\begin{equation}
C_b = \frac{1}{m_b}(X_P(b) - \bar{x}_b)(X_P(b) - \bar{x}_b)^T
\label{eq:ipca_cb}
\end{equation}
\noindent\textit{where (Eq.~\ref{eq:ipca_cb})}  
$X_P(b)$ represents the $b^{\text{th}}$ batch of the polynomially expanded feature matrix $X_P$ (as defined in Eq.~\eqref{eq:poly_matrix}),  
and $\bar{x}_b$ is the mean of that batch.

The eigen-decomposition step is then performed as:
\begin{equation}
S_b v_i = \lambda_i v_i, \quad i = 1, \ldots, k
\label{eq:ipca_eig}
\end{equation}
\noindent\textit{where (Eq.~\ref{eq:ipca_eig})}  
$\lambda_i$ and $v_i$ denote the $i^{\text{th}}$ eigenvalue and eigenvector of the updated covariance matrix $S_b$, respectively.

Finally, the projection of data onto the principal component subspace is given by:
\begin{equation}
T = X_P V_k, \quad V_k = [v_1, v_2, \ldots, v_k]
\label{eq:ipca_proj}
\end{equation}
\noindent\textit{where (Eq.~\ref{eq:ipca_proj})}  
$T$ is the matrix of principal component scores,  
and $V_k$ is the projection matrix formed by the top $k$ eigenvectors.  

Applying IPCA after cluster-wise polynomial expansion is novel because it enables incremental, memory efficient computation of principal components on nonlinearly transformed features, which is particularly advantageous for streaming or large-scale datasets.

\noindent\textbf{d) Standardization of Principal Components:}  
To ensure comparable scaling across all extracted components, the projected features are standardized as follows:
\begin{equation}
Z = \frac{T}{\sigma_T}
\label{eq:Z_std}
\end{equation}

After standardization, each transformed component satisfies:
\begin{equation}
\bar{z}_i \approx 0, \quad \sigma_{z_i} \approx 1
\label{eq:z_mean_std}
\end{equation}

\noindent\textit{where (Eq.~\ref{eq:Z_std})} $T$ is the matrix of principal component scores from Eq.~\eqref{eq:ipca_proj}, $\sigma_T$ is the standard deviation across components, and $Z$ is the standardized feature matrix.  

\noindent\textit{where (Eq.~\ref{eq:z_mean_std})} $\bar{z}_i$ and $\sigma_{z_i}$ are the mean and standard deviation of the $i^{\text{th}}$ standardized component.



Standardizing principal components after IPCA is a novel step because it ensures uniform scaling, improves interpretability, and enhances numerical stability which it is not addressed in traditional IPCA formulations.

\noindent\textbf{e) Final Compact Form:}  
The complete CPE-IPCA transformation is expressed as:

\begin{equation}
Z_{\text{final}} = \operatorname{concat}_{k=1}^{K} \; \text{IPCA}\!\left( \Phi\!\left( X_{C_k} \right) \right)
\label{eq:final_cpe_ipca}
\end{equation}

where, $Z_{\text{final}}$ in Eq.~\eqref{eq:final_cpe_ipca} is the final reduced feature matrix after applying the complete hierarchical CPE-IPCA pipeline,  
$\Phi(X_{C_k})$ denotes the polynomially expanded feature set of cluster $k$ (as defined in Eq.~\eqref{eq:poly_expansion} and Eq.~\eqref{eq:poly_matrix}),  
and $\mathrm{IPCA}(\Phi(X_{C_k}))$ represents the incremental PCA transformation applied to the polynomially expanded cluster features (as described in Eqs.~\eqref{eq:ipca_mean}--\eqref{eq:ipca_proj}).

This hierarchical formulation in Eq.~\eqref{eq:final_cpe_ipca} is novel as it unifies correlation-based clustering (Eqs.~\eqref{eq:cluster_def}--\eqref{eq:cluster_matrix}), polynomial expansion (Eqs.~\eqref{eq:poly_expansion}--\eqref{eq:poly_matrix}), incremental PCA (Eqs.~\eqref{eq:ipca_mean}--\eqref{eq:ipca_proj}), and post-standardization (Eqs.~\eqref{eq:Z_std}--\eqref{eq:z_mean_std}) into a coherent framework for nonlinear dimensionality reduction.


\noindent\textbf{Computation Time of the Proposed CPE-IPCA Method}

The total computation time of the proposed CPE-IPCA framework can be expressed as
\begin{equation}
T_{\mathrm{CPE\text{-}IPCA}} = \mathcal{O}(m n^2) + \mathcal{O}(m K^4 + B K^6),
\label{eq:cpeipca_full_time}
\end{equation}
where $m$ is the number of samples, $n$ is the number of original features, $K$ denotes the number of clusters obtained from correlation-based clustering (Eq.~\eqref{eq:cluster_matrix}), and $B$ is the number of batches used in Incremental PCA (Eqs.~\eqref{eq:ipca_mean}--\eqref{eq:ipca_proj}). 

In practical applications, $K \ll n$ and $B \ll m$, so the first term dominates, allowing the computation time to be approximated as
\begin{equation}
T_{\mathrm{CPE\text{-}IPCA}} \approx \mathcal{O}(m n^2),
\label{eq:cpeipca_simplified_time}
\end{equation}

resulting in a much faster execution compared to conventional nonlinear dimensionality reduction methods. For instance, Kernel PCA \((\mathcal{O}(m^3))\)~\cite{IEEE_ICIP_2010}, Isomap \((\mathcal{O}(m^3))\)~\cite{Rehman2022}, and t-SNE \((\mathcal{O}(m^2))\)~\cite{Pezzotti2017} scale quadratically or cubically with the number of samples, making them impractical for large-scale, high-dimensional datasets such as those derived from UWB signals.

For high-dimensional datasets where $n \ll m$, the computational cost of the proposed CPE-IPCA satisfies


\begin{equation}
T_{\mathrm{CPE\text{-}IPCA}} \approx \mathcal{O}(m n^2) \ll \mathcal{O}(m^2)
\end{equation}
\begin{equation}
T_{\mathrm{CPE\text{-}IPCA}} \approx \mathcal{O}(m n^2) \ll \mathcal{O}(m^3)
\end{equation}
demonstrating the scalability and efficiency of the proposed CPE-IPCA method.

Our proposed CPE-IPCA framework introduces a novel hierarchical approach that eliminates redundant features using correlation-based clustering, models non-linear dependencies via polynomial expansion, and produces compact representations through IPCA, providing a scalable and memory-efficient solution for large scale nonlinear datasets such as UWB.

\subsubsection{Post-PCA Standardization Approach (PPSA) Overview} 

Our proposed PPSA is a novel enhancement of traditional PCA for high-dimensional UWB data. It performs post-projection standardization to produce uniformly scaled features, improving interpretability, numerical stability, and comparability across components. The full procedure is summarized in Algorithm~\ref{alg:pps-summary}.

\begin{algorithm}[h]
\caption{Post-PCA Standardization Approach (PPSA)}
\label{alg:pps-summary}
\begin{algorithmic}[1]
\REQUIRE Feature matrix $X \in \mathbb{R}^{m \times n}$, number of PCA components $r$
\ENSURE Standardized reduced feature matrix $X' \in \mathbb{R}^{m \times r}$

\STATE Compute the mean vector $\mu$ of $X$
\STATE Center the data by subtracting $\mu$ from each feature
\STATE Compute PCA eigenvectors and select top $r$ components
\STATE Project the centered data onto the selected PCA components
\STATE Standardize each projected component to zero mean and unit variance
\STATE Return the standardized reduced feature matrix $X'$
\end{algorithmic}
\end{algorithm}

\textbf{PPSA Pipeline Equations:}

\textbf{(a) Data Centering:}  
To eliminate mean offset and prepare the data for projection, the original data matrix is centered as  

\begin{equation} \label{eq:center}
X_{\mathrm{centered}} = X - \mathbf{1}_m \mu^T
\end{equation}

where Eq.~\eqref{eq:center}:  
\(X \in \mathbb{R}^{m \times n}\) (original data matrix containing \(m\) samples and \(n\) features)  
\(\mu \in \mathbb{R}^{n}\) (mean vector of each feature computed across all samples)  
\(\mathbf{1}_m \in \mathbb{R}^{m}\) (column vector of ones used to replicate the feature means across all rows)  
\(X_{\mathrm{centered}}\) (mean-centered data obtained by subtracting the mean vector from each sample)  

\textbf{(b) PCA Projection:}  
The centered data from Eq.~\eqref{eq:center} is then projected onto the principal subspace spanned by the top \(r\) eigenvectors as  

\begin{equation} \label{eq:pca_projection}
X_{\mathrm{projected}} = X_{\mathrm{centered}} V
\end{equation}

where Eq.~\eqref{eq:pca_projection}:  
\(V \in \mathbb{R}^{n \times r}\) (matrix containing the top \(r\) eigenvectors of the covariance matrix of \(X_{\mathrm{centered}}\))  
\(X_{\mathrm{projected}}\) (data projected onto the reduced PCA subspace using the eigenvector matrix \(V\))  

\textbf{(c) Post-PCA Standardization (Novelty):}  
To ensure equal contribution of each principal component and prevent dominance by high-variance components, the PCA-projected data from Eq.~\eqref{eq:pca_projection} is standardized column-wise.  
The complete transformation can be expressed as  

\begin{equation} \label{eq:post_pca_std_overall}
X' = \sigma(X_{\mathrm{projected}}) = \sigma\big((X - \mathbf{1}_m \mu^T)V\big)
\end{equation}

where Eq.~\eqref{eq:post_pca_std_overall}:  
\(\sigma(X_{\mathrm{projected}})\) (column-wise standardization applied to the PCA-projected data to scale each feature to zero mean and unit variance)  

\(X'\) (standardized PCA-projected data representing the combination of centering, projection, and normalization in a single transformation)

The key novelty of PPSA lies in the post-PCA standardization step (Eq.~\eqref{eq:post_pca_std_overall}), which is particularly useful for non-linear data, ensuring that all principal components contribute equally, improving interpretability, enhancing stability, and facilitating reliable downstream analysis.


\textbf{(d) Computational Complexity}

The computational cost of PPSA is expressed as  
\begin{equation} \label{eq:ppsa_total}
T_{\mathrm{PPSA}} = T_{\mathrm{Centering}} + T_{\mathrm{PCA}} + T_{\mathrm{Standardization}}
\end{equation}
\begin{equation} \label{eq:ppsa_order}
T_{\mathrm{PPSA}} = O(m n) + O(\min(m n^2, m^2 n)) + O(m r)
\end{equation}

The \(\min(m n^2, m^2 n)\) term arises because the PCA computational cost depends on the relative sizes of \(m\) and \(n\), choosing the more efficient computation path.  

Since \(m \gg n\) and \(r \ll n\), the standardization term \(O(m r)\) is negligible, and among the remaining terms, \(O(\min(m n^2, m^2 n)) = O(m n^2)\) dominates. Therefore, the overall complexity simplifies to  

\begin{equation} \label{eq:ppsa_simplified}
T_{\mathrm{PPSA}} \approx O(m n^2)
\end{equation}

where Eqs.~\eqref{eq:ppsa_total}--\eqref{eq:ppsa_simplified}:  
\(m\) (number of samples, with \(m \gg n\)),  
\(n\) (number of original features), and  
\(r \ll n\) (number of PCA components used in the dimensionality reduction)

\subsection{Pattern Mining via Association Rule Mining}

To identify frequent dependencies among features within the reduced-dimensional data, we employed \textit{Association Rule Mining (ARM)}~\cite{alaghbari2024association}. We selected ARM for its interpretability and its ability to uncover hidden feature relationships that may not emerge through dimensionality reduction alone. We applied two algorithms: \textit{Apriori}~\cite{alaghbari2024association}, which iteratively expands frequent itemsets, and \textit{FP-Growth}~\cite{thilina}, which enhances efficiency by constructing an FP-tree to eliminate repeated candidate generation.

Each rule is represented as \(A \Rightarrow C\), where \(A\) (antecedent) and \(C\) (consequent) denote subsets of co-occurring features. We evaluated the significance of each rule using three standard metrics, \textit{Support}, \textit{Confidence}, and \textit{Lift} defined as:

\begin{equation} \label{eq:support}
\mathrm{Support}(A \Rightarrow C) = \frac{|A \cup C|}{N}
\end{equation}
where \(N\) is the total number of transactions, and \(|A \cup C|\) is the count containing both \(A\) and \(C\).

\begin{equation} \label{eq:confidence}
\mathrm{Confidence}(A \Rightarrow C) = \frac{\mathrm{Support}(A \cup C)}{\mathrm{Support}(A)}
\end{equation}
measures the probability of \(C\) occurring when \(A\) occurs.

\begin{equation} \label{eq:lift}
\mathrm{Lift}(A \Rightarrow C) = \frac{\mathrm{Confidence}(A \Rightarrow C)}{\mathrm{Support}(C)}
\end{equation}
indicates the correlation strength between \(A\) and \(C\) beyond random chance.

Support (Eq.~\ref{eq:support}) ensures statistical relevance, Confidence (Eq.~\ref{eq:confidence}) captures predictive strength, and Lift (Eq.~\ref{eq:lift}) validates non-random associations. Rules exceeding the defined thresholds were retained, revealing discriminative and interpretable feature relationships essential for subsequent classification.

\subsection{Human Activity Classification}

In the final phase of our framework, we focused on recognizing human activities using two classifiers: \textit{Random Forest (RF)}~\cite{thakur2023grrf} and the \textit{Vector Space Model (VSM)}~\cite{sunori2024knn}. We applied these models to feature sets refined through dimensionality reduction and pattern discovery. To maintain class balance and avoid bias, we stratified the dataset and partitioned it into 70\% training and 30\% testing subsets.

We evaluated classification performance using four standard metrics: \textit{Precision}, \textit{Recall}, \textit{F1-score}, and \textit{Accuracy}~\cite{noori2021uwb}, defined as follows:

\begin{equation} \label{eq:precision_recall}
\mathrm{Precision} = \frac{TP}{TP + FP}, \quad
\mathrm{Recall} = \frac{TP}{TP + FN}
\end{equation}

\begin{equation} \label{eq:f1score}
F1 = 2 \times \frac{\mathrm{Precision} \times \mathrm{Recall}}{\mathrm{Precision} + \mathrm{Recall}}
\end{equation}

\begin{equation} \label{eq:accuracy}
\mathrm{Accuracy} = \frac{TP + TN}{TP + FP + FN + TN}
\end{equation}

In Equations~\eqref{eq:precision_recall}–\eqref{eq:accuracy}, \(TP\), \(TN\), \(FP\), and \(FN\) represent the numbers of true positives, true negatives, false positives, and false negatives, respectively. {Precision measures the proportion of correctly identified activities among all predicted positives, while Recall quantifies the proportion of correctly detected activities among all actual positives. The F1-score  (Eq.~\eqref{eq:f1score}) provides the harmonic mean of Precision and Recall, and \textit{Accuracy} (Eq.~\eqref{eq:accuracy}) represents the overall proportion of correct classifications across all activity classes.

By integrating dimensionality reduction, pattern mining, and classification, we achieved robust and computationally efficient Human Activity Recognition (HAR) performance across diverse activity scenarios.

Experiments were conducted using our simulation-based framework, the novel UWB-PRISM-C (Ultra-Wideband Processing, Reduction, Integrated Simulation for Mining and Classification), developed to model and evaluate the complete Human Activity Recognition (HAR) pipeline from UWB data. UWB-PRISM-C integrates multiple analytical stages, including preprocessing with matrix correlation analysis to eliminate redundant features, non-linear dimensionality reduction through CPE-IPCA and PPSA, pattern mining via Apriori and FP-Growth, and predictive classification using Random Forest and the Vector Space Model (VSM). This unified design enables robust, interpretable, and scalable activity recognition performance.

The framework was implemented in Python 3.10, utilizing essential libraries such as NumPy, Pandas, SciPy, scikit-learn, Seaborn, Matplotlib, and mlxtend. All simulations were executed on a workstation equipped with an AMD Ryzen 7 3700X CPU, 14 GB RAM, and Ubuntu 20.04 operating system. To efficiently process the high-dimensional UWB data, chunked computation and sparse matrix structures were employed, ensuring memory-efficient, scalable, and reproducible experimental performance.

\section{Experimental Results}
This section presents the experimental evaluation of the proposed UWB-PRISM-C framework. The analysis begins with an examination of the dataset characteristics and linearity assessment, followed by evaluations of dimensionality reduction, pattern discovery, and classification performance. The results demonstrate the effectiveness and robustness of the proposed approach across all stages of the Human Activity Recognition (HAR) pipeline.

\subsection{Data Characterization and Linearity Analysis}

The UWB dataset comprised 17,063 observations and 2,040 fully numeric, standardized features, obtained by separating each complex CIR entry into real and imaginary parts. Low contribution attributes, such as \textit{RX Power} and \textit{Accumulated FP Index}, were removed to enhance feature relevance.

Linearity analysis using Linear Regression ($R^2=0.1331$) versus Random Forest regression ($R^2=0.8677$) revealed that the dataset is predominantly non-linear, indicating complex underlying relationships among features.

\subsection{Proposed Dimensionality Reduction Techniques}
This section presents the results of PCA, Clustered Polynomial Expansion with Incremental PCA (CPE-IPCA), and Post-PCA Standardization, applied to a non-linear, high dimensional dataset.
\subsubsection{Traditional Principal Component Analysis}
All features were first standardized to zero mean and unit variance before applying PCA to reduce dimensionality while retaining maximal variance. Figure~\ref{fig:2} shows the individual and cumulative explained variance. The first principal component captures about 5\% of the total variance, while 723 components are required to reach 90\%, indicating high feature redundancy and complex inter-feature correlations.

\subsubsection{Clustered Polynomial Expansion with Incremental PCA}

CPE-IPCA was applied to the standardized nonlinear high-dimensional dataset. Table~\ref{tab:variance} shows that PC1 accounts for 12.98\% of the variance. The Knee-Locator identified PC13 as the elbow (86\% cumulative variance), whereas the Difference Threshold method indicated PC1. All 50 components retained 100\% of the variance, confirming that CPE-IPCA effectively compresses the dataset while preserving information. Figure~\ref{fig:3} visualizes the explained and cumulative variance trends. The reduced components were centered around zero and scaled to unit variance, ensuring uniform contribution across features. 

\subsubsection{Post-PCA Standardization} 
PPSA enhances both interpretability and consistency, while retaining nearly all of the original variance. As shown in Table~\ref{tab:3} and Figure~\ref{fig:4}, the first two principal components (PCs) account for 34.72\% of the total variance, and the first ten PCs explain 76.14\%. Dimensionality was reduced to 80 components, preserving 99.1\% of the total variance. The data were centered (with means approximately zero) and scaled to unit variance, ensuring a balanced contribution across all features.

\begin{table}[!h]
\centering
\caption{Explained and Cumulative Variance by CPE-IPCA (Top 50 Components)}
\label{tab:variance}
\begin{tabular}{|c|c|c|}
\hline
\textbf{Component} & \textbf{Explained Variance (\%)} & \textbf{Cumulative Variance (\%)} \\
\hline
PC1 & 12.98 & 12.98 \\
PC2 & 10.34 & 23.32 \\
PC3 & 9.02  & 32.34 \\
PC4 & 8.51  & 40.85 \\
PC5 & 7.43  & 48.28 \\
PC6 & 6.81  & 55.09 \\
PC7 & 5.73  & 60.82 \\
PC8 & 5.04  & 65.86 \\
PC9 & 4.23  & 70.09 \\
PC10 & 3.85 & 73.94 \\
PC11 & 3.61 & 77.55 \\
PC12 & 3.72 & 81.27 \\
PC13 & 4.63 & 85.90 \\
PC14 & 3.07 & 88.97 \\
PC15 & 2.55 & 91.52 \\
PC16 & 2.36 & 93.88 \\
PC17 & 2.19 & 96.07 \\
PC18 & 1.78 & 97.85 \\
PC19 & 1.17 & 99.02 \\
PC20 & 0.78 & 99.80 \\
PC21--PC50 & 0.20 & 100.00 \\
\hline
\end{tabular}
\end{table}

\begin{table}[!h]
\caption{Explained and Cumulative Variance: Post-PCA Standardization}
\label{tab:3}
\centering
\begin{tabular}{|c|c|c|}
\hline
\textbf{Component} & \textbf{Explained Variance (\%)} & \textbf{Cumulative Variance (\%)} \\
\hline
PC1  & 17.48 & 17.48 \\
PC2  & 17.24 & 34.72 \\
PC3  & 9.61  & 44.33 \\
PC4  & 9.46  & 53.79 \\
PC5  & 4.66  & 58.45 \\
PC6  & 4.55  & 63.00 \\
PC7  & 3.70  & 66.70 \\
PC8  & 3.62  & 70.32 \\
PC9  & 2.96  & 73.28 \\
PC10 & 2.86  & 76.14 \\
PC11 & 1.95  & 78.09 \\
PC12 & 1.88  & 79.97 \\
PC13 & 1.62  & 81.59 \\
PC14 & 1.57  & 83.16 \\
PC15 & 1.51  & 84.67 \\
PC16 & 1.48  & 86.15 \\
PC17 & 1.45  & 87.60 \\
PC18 & 1.41  & 89.01 \\
PC19 & 1.39  & 90.40 \\
PC20 & 1.38  & 91.78 \\
PC21--PC80 & 7.99 & 99.07.00 \\
\hline
\end{tabular}
\end{table}

\begin{figure*}[t]
    \centering
    \includegraphics[width=0.9\textwidth]{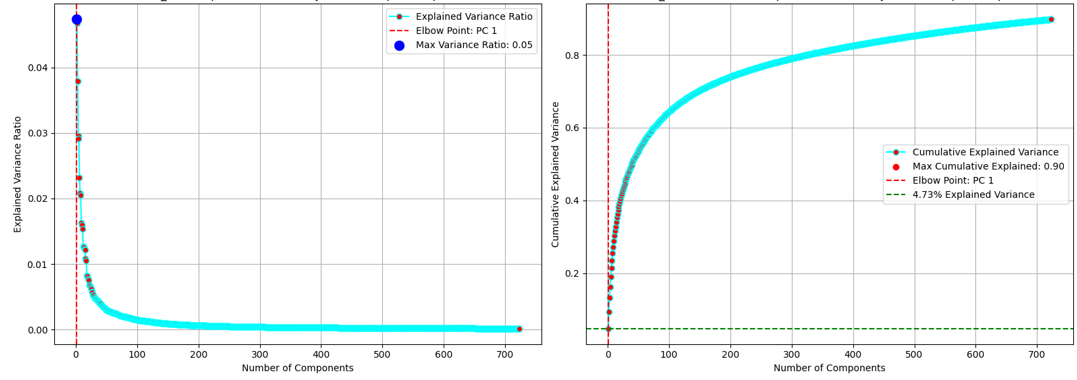}
    \put(-368,170){\textbf{PCA:Explained and Cumulative Explained Variance }}
    \caption{Explained variance (left) and cumulative explained variance (right) for each principal component after PCA. PC1 captures 4.73\% variance; 723 components are needed to retain 90\% of the variance.}
    \label{fig:2}
\end{figure*}

\begin{figure*}[t]
    \centering
    \includegraphics[width=0.9\textwidth]{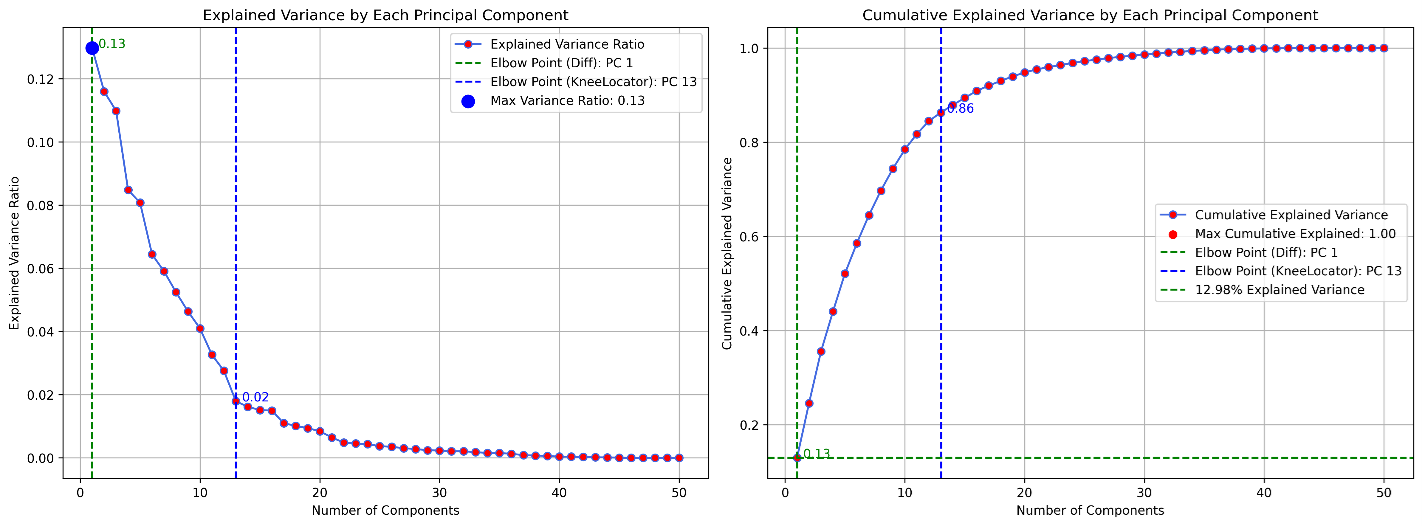}
    \put(-368,174){\textbf{CPE-IPCA: Explained and Cumulative Explained Variance}}
    \caption{The CPE-IPCA analysis identified explained and cumulative explained variances for each principal component. PC1 explains 13\% of the variance, while PC13 marks the elbow with 86\% cumulative variance.}
    \label{fig:3}
\end{figure*}

\begin{figure*}[!t]
    \centering
    \includegraphics[width=\textwidth]{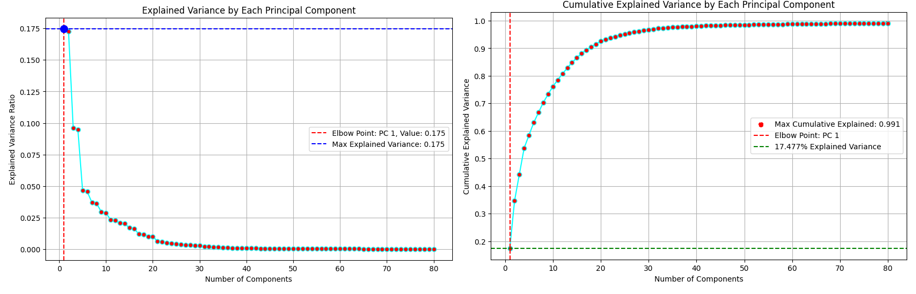}
    \put(-368,182){\textbf{PPSA: Explained and Cumulative Explained Variance}}
    \caption{PPSA: The left plot shows the individual explained variance for each principal component, highlighting PC1 with the highest contribution of 17.5\%. The right plot illustrates the cumulative explained variance, which reaches 99.1\%.}
    \label{fig:4}
\end{figure*}

\subsection{Association Rule Mining Applied to the Post-PCA Dataset}
The Apriori and FP-Growth algorithms were applied with a minimum support of 0.03 to a binary-transformed Post-PCA dataset containing 9,913 rows. Apriori identified the 20 most frequent itemsets, with support values ranging from 0.291 to 0.356; [PC1], [PC2], [PC4], and [PC3] had the highest supports. FP-Growth produced nearly identical results but completed the analysis more quickly due to its tree-based structure.

\subsubsection{Frequent Rule Patterns and Key Components}
As shown in Figures~\ref{fig:5} and~\ref{fig:6}, the top rules by \textit{lift} involved PC43,PC32,and PC7 (lift = 2.38), while PC1 had the {highest support} (0.15), primarily associated with the “noactivity” class. These components are key indicators of activity patterns and provide valuable targets for classification or feature selection.

\subsubsection{Rule Counts and Metrics by Thresholds (Post-PCA)}

Table~\ref{tab:rulesummary} presents the impact of varying support and confidence thresholds on the number and quality of association rules extracted from the Post-PCA dataset. Lower confidence thresholds (e.g., 0.1) generate a large number of rules ($\approx 39{,}000$), whereas higher thresholds (0.5) produce fewer but more reliable rules (26), illustrating the classical precision–recall trade-off. This analysis highlights how threshold selection influences both the quantity and statistical strength of discovered patterns.

\begin{table}[htbp]
\caption{Post-PCA: Summary of Rule Counts and Key Metrics}
\label{tab:rulesummary}
\centering
\begin{tabular}{|l|c|c|c|}
\hline
\textbf{Rule Set} & \textbf{Rule Count} & \textbf{Mean} & \textbf{Std Dev} \\
\hline
Confidence 0.1      & 39,851$\times$10 & 0.254 & 0.110 \\
Confidence 0.3      & 16,340$\times$10 & 0.359 & 0.045 \\
Confidence 0.5      & 26$\times$10     & 0.516 & 0.018 \\
Lift (all)          & 45,150$\times$10 & 1.086 & 0.122 \\
Support 0.1         & 376$\times$10    & 0.112 & 0.011 \\
Support 0.3–0.5     & 0               & —     & —     \\
\hline
\end{tabular}
\end{table}

\subsubsection{Network Visualization of Apriori and FP-Growth}
The network visualization in Figure~\ref{fig:7} (Apriori: top, FP-Growth: bottom) contains 2,892 nodes and 45,150 edges. Although both algorithms generated the same set of rules, their search strategies differ: Apriori uses breadth-first, while FP-Growth follows depth-first, leading to distinct visual layouts. Both graphs show strongly connected regions and clear patterns, confirming the consistency of the extracted rules.  

Edge colors represent lift strength, highlighting the most influential associations. The FP-Growth graph demonstrates faster detection of key patterns due to its compact tree-based structure. Overall, these visualizations provide an intuitive overview of rule interconnections and their relative importance, aiding interpretation and decision-making.
\subsection{Association Rule Mining Applied to CPE-IPCA Dataset}
Apriori and FP-Growth were applied to the binary CPE-IPCA dataset (10,733 records, min support = 0.0181); the top 20 itemsets had support 0.239–0.272, highest for [PC50] (0.272), [PC39]/[PC33] (0.270), [PC19] (0.267), showing balanced feature representation, while FP-Growth gave nearly identical results faster.


\begin{figure}[t]
    \centering
    \includegraphics[width=0.48\textwidth]{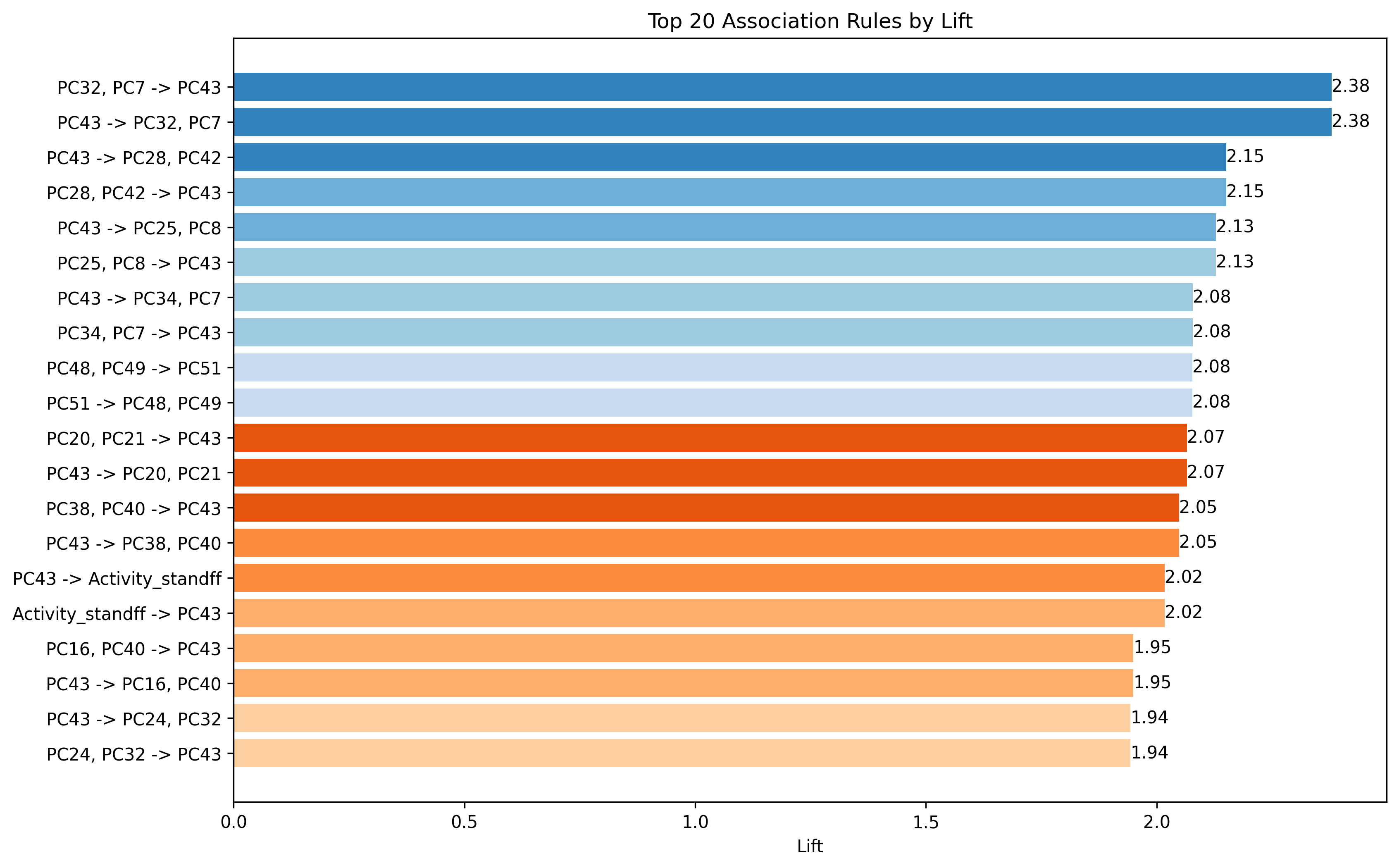}
     \put(-200,160){\textbf{PPSA:Top 20 Association Rules by Lift}}
    \caption{Top 20 association rules ranked by lift.}
    \label{fig:5}
\end{figure}
\begin{figure}[H]
    \centering
    \includegraphics[width=0.48\textwidth]{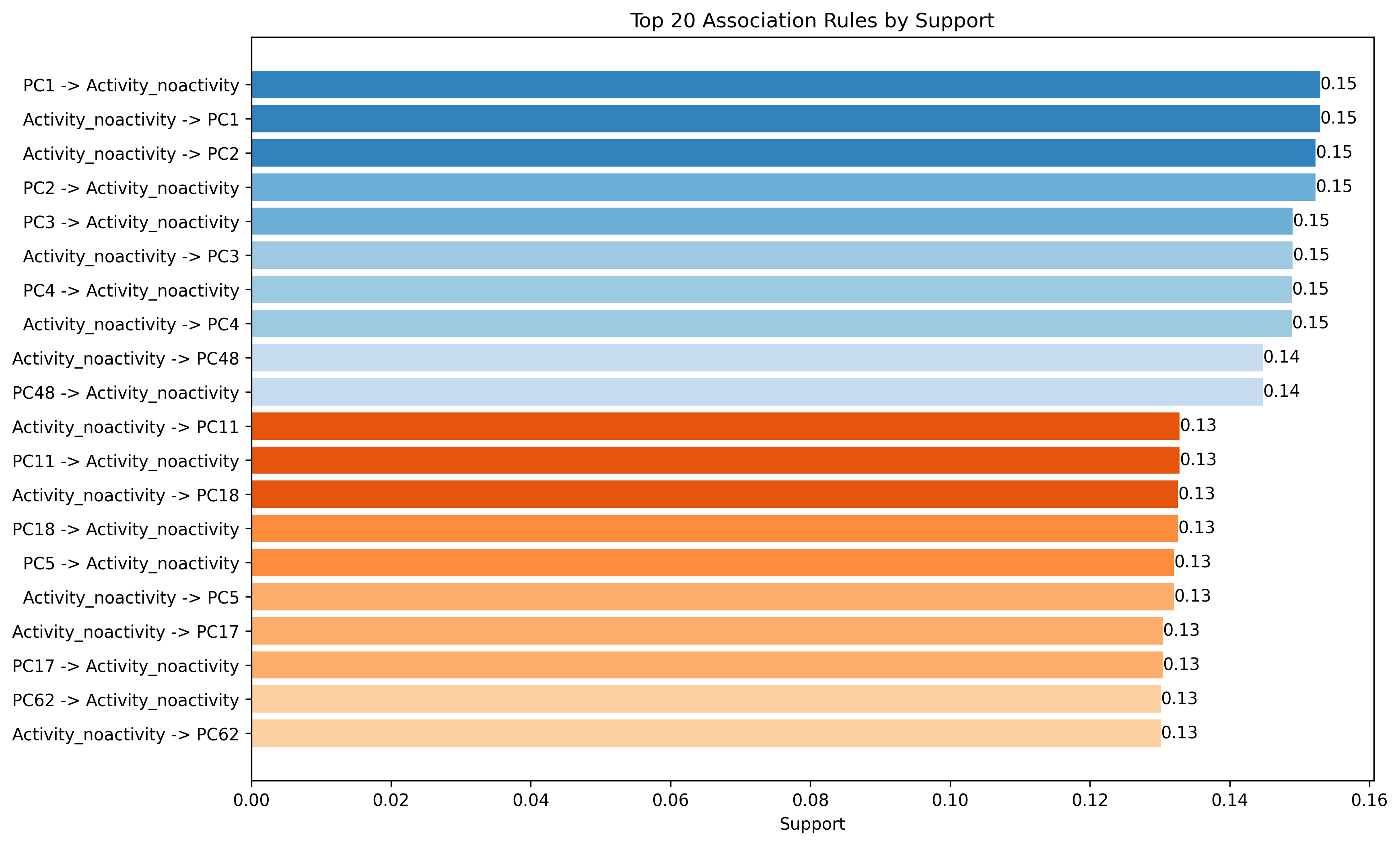}
     \put(-200,160){\textbf{PPSA:Top 20 Association Rules by Support}}
    \caption{Top 20 association rules ranked by Support.}
    \label{fig:6}
\end{figure}
\FloatBarrier

\begin{figure}[t]
    \centering

    \begin{overpic}[width=0.5\textwidth]{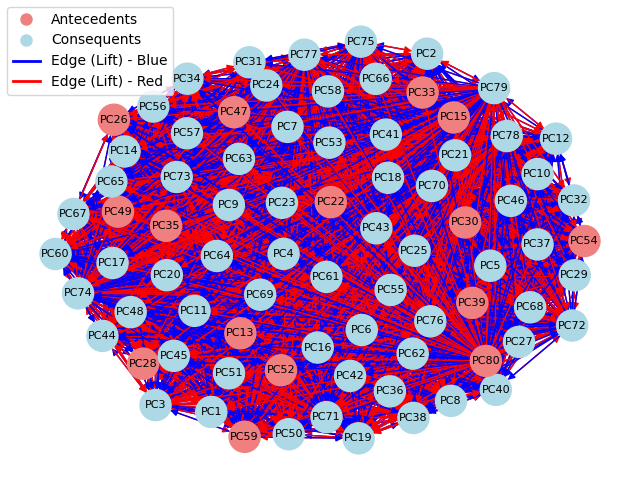}
        \put(0,105){\textbf{(a) Apriori: Network Visualization of Association Rules}}
    \end{overpic}
    \vspace{0.012em} 

    \begin{overpic}[width=0.5\textwidth]{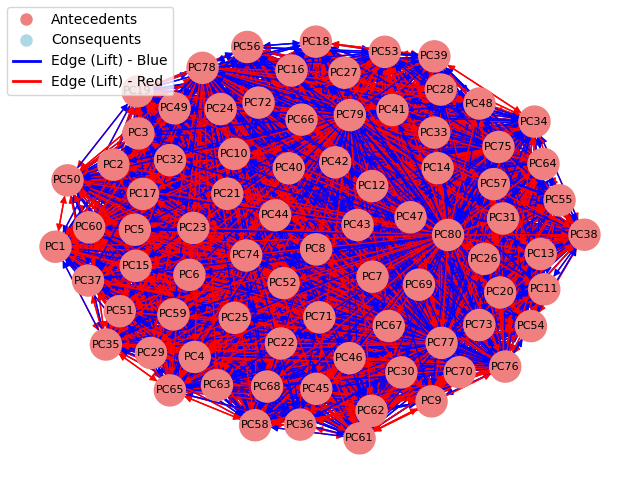}
        \put(0,158){\textbf{PPSA: Network Visualization of Association Rules}}
    \end{overpic}

    \caption{PPSA: Network graphs from Apriori (top) and FP-Growth (bottom) showing principal component associations. Edge colors indicate lift strength; FP-Growth offers faster detection.}
    \label{fig:7}
\end{figure}

\subsubsection{CPE-IPCA Key Rules and Feature Correlations}
As shown in Figures~\ref{fig:8} and \ref{fig:9}, strong associations and key features were identified. The rule {PC47, PC9~$\rightarrow$~PC1, PC48, PC7} achieved the highest lift (9.35), indicating strong dependencies. Several other rules had lifts above 8.4, confirming robust feature relationships. The {noactivity}~$\rightarrow$~PC50 rule showed the highest support (0.12), while other {noactivity}-related rules with PCs 39, 12, and 15 had supports around 0.11, highlighting their significance in activity classification.

\begin{figure}[h]
    \centering
    \includegraphics[width=0.48\textwidth]{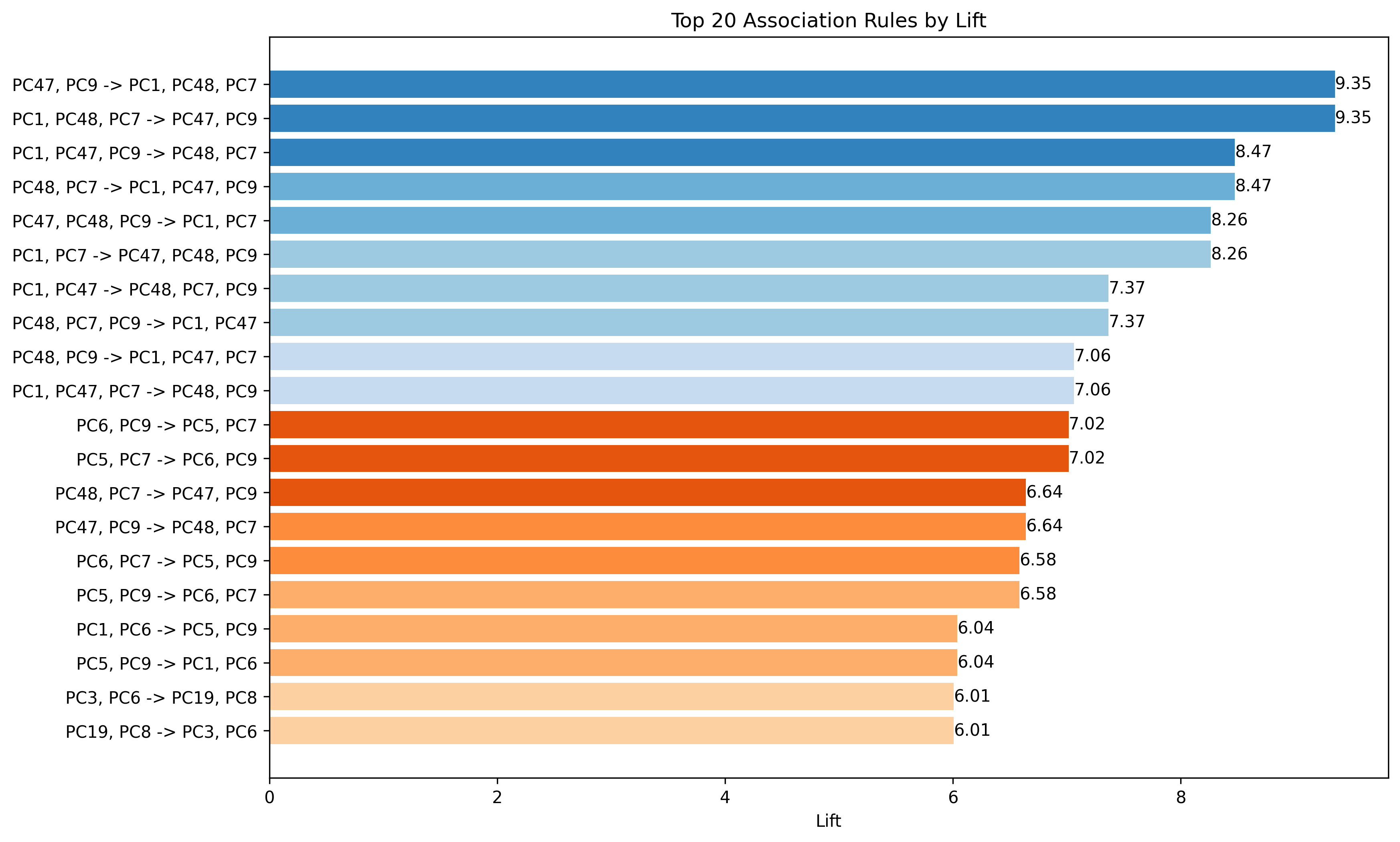}
     \put(-200,155){\textbf{CPE-IPCA:Top 20 Association Rules by Lift}}
    \caption{Top 20 association rules ranked by lift.}
    \label{fig:8}
\end{figure}
\begin{figure}[h]
    \centering
    \includegraphics[width=0.48\textwidth]{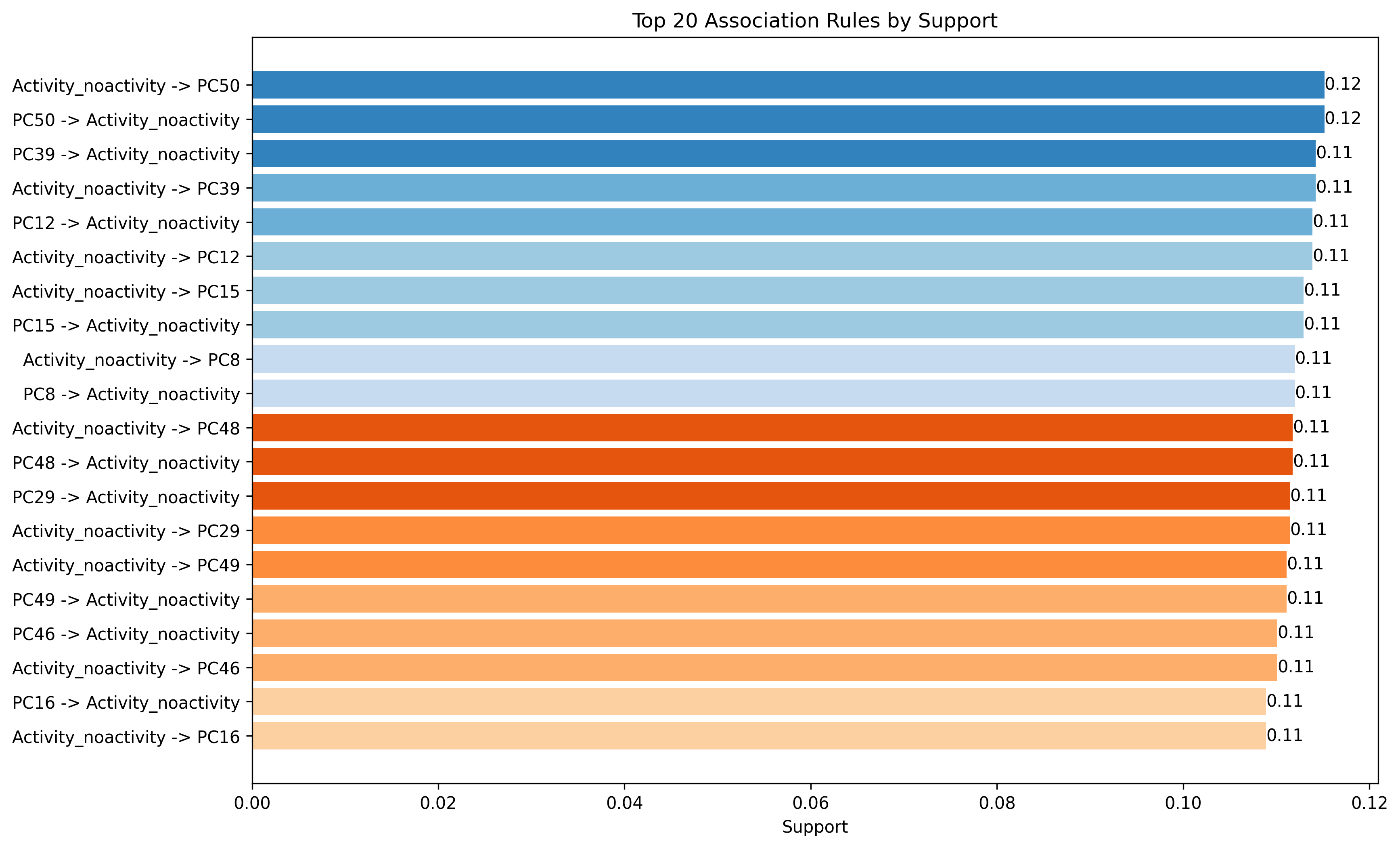}
     \put(-200,155){\textbf{CPE-IPCA:Top 20 Association Rules by Support}}
    \caption{Top 20 association rules ranked by Support.}
    \label{fig:9}
\end{figure}

\subsubsection{Rule Extraction and Threshold Effects} 
Table~\ref{tab:rulesummary_cpeipca} shows that using support = 0.05, confidence = 0.2, and lift = 0.4 balances rule reliability and coverage. Increasing confidence from 0.05 to 0.4 raised average confidence (0.22→0.46) while lift stayed at 1.42. Meaningful rules appeared only at support 0.05 (mean 0.07), showing low support is needed for sparse data.
\subsubsection{Network Structure and Algorithmic Differences} 
As shown in Figure~\ref{fig:10} the rule network had 1,695 nodes and 59,166 edges, indicating dense interconnections. While Apriori and FP-Growth produced similar networks, their different search strategies, breadth, first vs depth-first, led to structural differences.

\begin{table}[!t]
\caption{CPE-IPCA: Rule Summary Across Thresholds}
\label{tab:rulesummary_cpeipca}
\centering
\begin{tabular}{|l|c|c|c|}
\hline
\textbf{Filtering Type} & \textbf{Threshold} & \textbf{Rule Set Size} & \textbf{Mean Value} \\
\hline
Confidence-based & 0.05 & 59,022 & 0.22 \\
Confidence-based & 0.2  & 30,797 & 0.33 \\
Confidence-based & 0.4  & 4,770  & 0.46 \\
Lift (all)       & 0.05, 0.2, 0.4 & 59,166 & 1.42 \\
Support-based    & 0.05 & 2,218  & 0.07 \\
Support-based    & 0.2  & 0      & —    \\
\hline
\end{tabular}
\end{table}

\begin{figure}[t]
    \centering
    \begin{overpic}[width=0.5\textwidth]{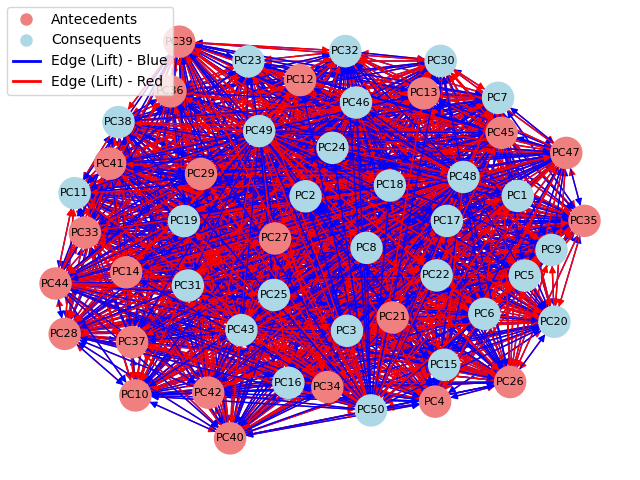}
        \put(0,155){\textbf{ Apriori: Network Visualization of Association Rules}}
    \end{overpic}
    \vspace{0.012em} 

    \begin{overpic}[width=0.5\textwidth]{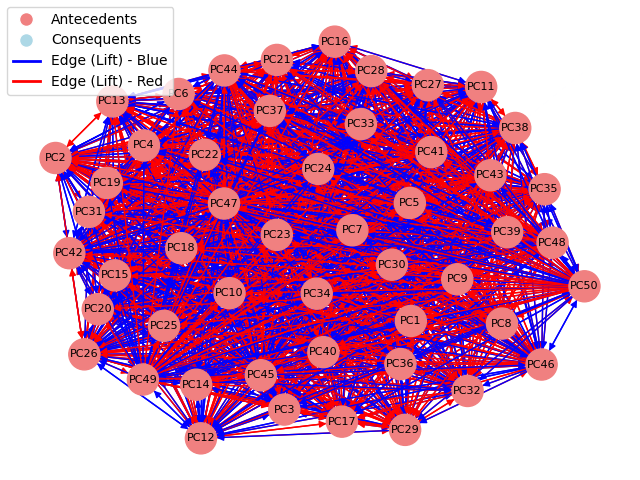}
        \put(0,155){\textbf{CPE-IPCA: Network Visualization of Association Rules}}
    \end{overpic}

    \caption{CPE-IPCA: Network graphs from Apriori (top) and FP-Growth (bottom) showing principal component associations. Edge colors denote lift strength; FP-Growth detects faster.}
    \label{fig:10}
    
\end{figure}

\subsection{Activity Classification Using Rule-Based Features}
This section evaluates the performance of Random Forest (RF) and Vector Space Model (VSM) on rule-based datasets from dimensionality-reduced representations. The CPE-IPCA method produced 59,166 samples, and PPSA 45,150 samples, each with 10 binary features, providing interpretable, high-level inputs for activity classification. These features capture strong associations between sensor measurements and activities, enhancing model interpretability, and experiments used standard training/testing splits to ensure unbiased evaluation.

\subsubsection{Activity Classification via Random Forest}
The Random Forest classifier achieved consistently high accuracy on both PPSA- and CPE-IPCA-based rule datasets. As shown in Table~\ref{tab:rf_performance_comparison}, most activity classes reached perfect or near-perfect accuracy. The confusion matrices (Tables~\ref{tab:confusion_matrix1} and \ref{tab:confusion_matrix2}) further confirm strong diagonal correctness.

\subsubsection{Activity Classification via Vector Space Model}
VSM achieved near-perfect results on rule-based datasets from PPSA and CPE-IPCA. On PPSA (45,150 × 10), it reached 100\% precision, recall, and F1 across all classes, including liedown, sit, and walk. On CPE-IPCA (59,166 × 10), recall slightly dropped to 0.99 for liedown and 0.98 for noactivity, with only 0.02 misclassification (Tables~\ref{tab:vsm_combined}, ~\ref {tab:confusion_ppsa} and \ref{tab:confusion_CPE-IPCA}).


\begin{table}[!t]
\caption{Random Forest on PPSA and CPE-IPCA rule datasets (30\% test) with Precision, Recall, F1, Support, accuracy, and class averages.}
\centering
\scriptsize
\setlength{\tabcolsep}{4pt}
\begin{tabular}{l|cccc|cccc}
\hline
\multirow{2}{*}{\textbf{Class}} & \multicolumn{4}{c|}{\textbf{PPSA}} & \multicolumn{4}{c}{\textbf{CPE-IPCA}} \\
\cline{2-9}
 & Prec. & Rec. & F1 & Sup. & Prec. & Rec. & F1 & Sup. \\ \hline
Standff & 1.00 & 1.00 & 1.00 & 11021 & 1.00 & 1.00 & 1.00 & 16558 \\
Liedown & 1.00 & 0.20 & 0.33 & 5 & 1.00 & 0.88 & 0.94 & 17 \\
Noactivity & 0.98 & 0.99 & 0.99 & 2460 & 0.97 & 0.97 & 0.97 & 1128 \\
Sit & 1.00 & 1.00 & 1.00 & 22 & 1.00 & 1.00 & 1.00 & 11 \\
Stand & 1.00 & 0.80 & 0.89 & 25 & 1.00 & 1.00 & 1.00 & 26 \\
Walk & 1.00 & 1.00 & 1.00 & 12 & 1.00 & 1.00 & 1.00 & 10 \\ \hline
Accuracy & \multicolumn{4}{c|}{0.99} & \multicolumn{4}{c}{1.00} \\
Macro Avg & 1.00 & 0.83 & 0.87 & 13545 & 1.00 & 0.97 & 0.98 & 17750 \\
Weighted Avg & 0.99 & 0.99 & 0.99 & 13545 & 1.00 & 1.00 & 1.00 & 17750 \\ \hline
\end{tabular}
\label{tab:rf_performance_comparison}
\end{table}

\vspace{1mm}

\begin{table}[H]
\caption{Random Forest Confusion Matrix on Activity Rule-Set Extracted from PPSA}
\centering
\setlength{\tabcolsep}{4pt}
\begin{tabular}{l|cccccc}
\hline
\textbf{Actual \textbackslash Predicted} & Standff & Liedown & Noactivity & Sit & Stand & Walk \\ \hline
Standff & 1.00 & 0.00 & 0.00 & 0.00 & 0.00 & 0.00 \\
Liedown & 0.80 & 0.20 & 0.00 & 0.00 & 0.00 & 0.00 \\
Noactivity & 0.01 & 0.00 & 0.99 & 0.00 & 0.00 & 0.00 \\
Sit & 0.00 & 0.00 & 0.00 & 1.00 & 0.00 & 0.00 \\
Stand & 0.20 & 0.00 & 0.00 & 0.00 & 0.80 & 0.00 \\
Walk & 0.00 & 0.00 & 0.00 & 0.00 & 0.00 & 1.00 \\ \hline
\end{tabular}
\label{tab:confusion_matrix1}
\end{table}


\begin{table}[H]
\caption{Random Forest Confusion Matrix on Activity Rule-Set Extracted from CPE-IPCA}
\centering
\setlength{\tabcolsep}{4pt}
\begin{tabular}{l|cccccc}
\hline
\textbf{Actual \textbackslash Predicted} & Standff & Liedown & Noactivity & Sit & Stand & Walk \\ \hline
Standff & 1.00 & 0.00 & 0.00 & 0.00 & 0.00 & 0.00 \\
Liedown & 0.06 & 0.88 & 0.06 & 0.00 & 0.00 & 0.00 \\
Noactivity & 0.03 & 0.00 & 0.97 & 0.00 & 0.00 & 0.00 \\
Sit & 0.00 & 0.00 & 0.00 & 1.00 & 0.00 & 0.00 \\
Stand & 0.00 & 0.00 & 0.00 & 0.00 & 1.00 & 0.00 \\
Walk & 0.00 & 0.00 & 0.00 & 0.00 & 0.00 & 1.00 \\ \hline
\end{tabular}
\label{tab:confusion_matrix2}
\end{table}

\vspace{-2mm}

\begin{table}[H]
\caption{VSM performance on PPSA and CPE-IPCA rule features (30\% test) with Precision, Recall, F1, Support, and overall accuracy per activity class.}
\centering
\scriptsize
\setlength{\tabcolsep}{4pt}
\begin{tabular}{l|cccc|cccc}
\hline
\multirow{2}{*}{\textbf{Class}} & \multicolumn{4}{c|}{\textbf{PPSA}} & \multicolumn{4}{c}{\textbf{CPE-IPCA}} \\
\cline{2-9}
 & Prec. & Rec. & F1 & Sup. & Prec. & Rec. & F1 & Sup. \\ \hline
Standff & 1.00 & 1.00 & 1.00 & 11094 & 1.00 & 1.00 & 1.00 & 16558 \\
Liedown & 1.00 & 1.00 & 1.00 & 9 & 1.00 & 0.99 & 1.00 & 17 \\
Noactivity & 1.00 & 1.00 & 1.00 & 2388 & 1.00 & 0.98 & 0.99 & 1128 \\
Sit & 1.00 & 1.00 & 1.00 & 18 & 1.00 & 1.00 & 1.00 & 11 \\
Stand & 1.00 & 1.00 & 1.00 & 27 & 1.00 & 1.00 & 1.00 & 26 \\
Walk & 1.00 & 1.00 & 1.00 & 9 & 1.00 & 1.00 & 1.00 & 10 \\ \hline
Accuracy & \multicolumn{4}{c|}{1.00} & \multicolumn{4}{c}{1.00} \\
Macro Avg & 1.00 & 1.00 & 1.00 & 13545 & 1.00 & 1.00 & 1.00 & 17750 \\
Weighted Avg & 1.00 & 1.00 & 1.00 & 13545 & 1.00 & 1.00 & 1.00 & 17750 \\ \hline
\end{tabular}
\label{tab:vsm_combined}
\end{table}

\vspace{-2mm}

\begin{table}[H]
\centering
\caption{VSM Confusion Matrix on Activity Rule-Set Extracted from PPSA}
\label{tab:confusion_ppsa}
\begin{tabular}{lcccccc}
\hline
\textbf{Actual$\backslash$Pred} & Standff & Liedown & Noactivity & Sit & Stand & Walk \\ \hline
Standff & 1 & 0 & 0 & 0 & 0 & 0 \\
Liedown & 0 & 1 & 0 & 0 & 0 & 0 \\
Noactivity & 0 & 0 & 1 & 0 & 0 & 0 \\
Sit & 0 & 0 & 0 & 1 & 0 & 0 \\
Stand & 0 & 0 & 0 & 0 & 1 & 0 \\
Walk & 0 & 0 & 0 & 0 & 0 & 1 \\ \hline
\end{tabular}
\end{table}

\vspace{-2mm}

\begin{table}[H]
\centering
\caption{VSM Confusion Matrix on Activity Rule-Set Extracted from CPE-IPCA}
\label{tab:confusion_CPE-IPCA}
\begin{tabular}{lcccccc}
\hline
\textbf{Actual$\backslash$Pred} & Standff & Liedown & Noactivity & Sit & Stand & Walk \\ \hline
Standff & 1 & 0 & 0 & 0 & 0 & 0 \\
Liedown & 0 & 1 & 0 & 0 & 0 & 0 \\
Noactivity & 0.02 & 0 & 0.98 & 0 & 0 & 0 \\
Sit & 0 & 0 & 0 & 1 & 0 & 0 \\
Stand & 0 & 0 & 0 & 0 & 1 & 0 \\
Walk & 0 & 0 & 0 & 0 & 0 & 1 \\ \hline
\end{tabular}
\end{table}


\section{Discussion}
This study demonstrates that rule based features offer an effective and interpretable framework for recognizing human activities from UWB signals. After dimensionality reduction using PPSA and CPE-IPCA, Apriori and FP-Growth produced compact binary features that significantly improved classification efficiency and accuracy.

\subsection{Implications of Data Structure and Linearity}

Preprocessing simplified and standardized the complex UWB signals, with variance filtering highlighting the most informative features. The large performance gap between Linear Regression and Random Forest confirms that the data exhibits strong nonlinear relationships, justifying the use of nonlinear dimensionality reduction methods such as CPE-IPCA. Separately, PPSA, as a linear reduction technique, provides a stable, zero centered, unit variance representation of the data, serving as a reliable linear baseline for feature analysis and activity classification.

\subsection{Impact of Dimensionality Reduction Techniques}


Traditional PCA often requires a large number of components to capture sufficient variance in nonlinear, high-dimensional settings (Fig.~\ref{fig:2}). In contrast, CPE-IPCA addresses this limitation by integrating correlation-based clustering, polynomial expansion, and Incremental PCA, achieving full variance retention with significantly fewer components (Table~\ref{tab:variance}) while maintaining computational efficiency suitable for real-time environments (Fig.~\ref{fig:3}).

PPSA further enhances PCA by producing zero-centered, unit-variance components (Table~\ref{tab:3}, Fig.~\ref{fig:4}), thereby improving numerical stability and interpretability. However, due to its linear formulation, PPSA is less effective at capturing nonlinear dependencies when compared to CPE-IPCA.




\subsection{Association Rule Mining: Interpretability and Insights}

Association rule mining provides a systematic framework to uncover meaningful relationships between principal components and activity labels, enabling interpretable feature analysis for human activity recognition (HAR). In the Post-PCA Standardization Approach (PPSA), post-PCA standardization balances contributions across all components, mitigating the dominance of high-variance features and enhancing interpretability. Using Apriori and FP-Growth, key components such as PC43, PC32, and PC7 were identified in high-lift rules (Figure~\ref{fig:5}), whereas PC1 dominated high-support rules (Figure~\ref{fig:6}). Network visualization (Figure~\ref{fig:7}) revealed coherent and structured relationships among components, confirming that PPSA preserves latent dependencies critical for accurate feature prioritization, with FP-Growth delivering faster yet equally informative results.

In contrast, Clustered Polynomial Expansion with Incremental PCA (CPE-IPCA) captures nonlinear, cluster-aware associations that are not evident in linear reductions. Frequent itemset analysis revealed high-lift rules such as {PC47, PC9} $\rightarrow$ {PC1, PC48, PC7} (Figures~\ref{fig:8} and \ref{fig:9}), indicating strong co-occurrence patterns, while PCs 50, 39, 33, and 19 were linked to inactivity states. Network visualization for CPE-IPCA (Figure~\ref{fig:10}) confirmed structural consistency, and Table~\ref{tab:rulesummary_cpeipca} illustrates the effects of thresholds on rule precision. Collectively, these results demonstrate that association rule mining—augmented by network visualization for PPSA and itemset analysis for CPE-IPCA—effectively identifies compact, discriminative, and interpretable features, supporting robust and efficient HAR classification.




\subsection{Discussion: Activity Classification Using Rule-Based Features}

The proposed rule based binary features enabled highly accurate and computationally efficient classification using only 10 features per sample. Random Forest achieved consistently strong performance across both the PPSA and CPE-IPCA derived datasets (Table~\ref{tab:rf_performance_comparison}), with the confusion matrices (Tables~\ref{tab:confusion_matrix1}, \ref{tab:confusion_matrix2}) indicating only a negligible number of misclassifications. Notably, the nonlinear structure captured by CPE-IPCA improved discrimination for low-support classes such as Liedown, demonstrating its effectiveness in preserving subtle, activity specific cues.

VSM likewise achieved near perfect accuracy (Tables~\ref{tab:vsm_combined}, \ref{tab:confusion_ppsa}, \ref{tab:confusion_CPE-IPCA}), benefiting from sparse, rule-based binary representations that support rapid and interpretable decision making. Collectively, these findings confirm that rule-based feature construction offers a robust, scalable, and highly accurate pathway for UWB-based HAR, reinforcing the practicality of the proposed framework for real-time and resource-constrained environments.

\subsection{Comparison with Existing Methods}

The experimental results confirm that our proposed Human Activity Recognition (HAR) framework, integrating the novel dimensionality reduction methods CPE-IPCA and Post-PCA Standardization (PPSA), achieves superior performance compared to existing approaches. The fall detection method by Wang \emph{et al.}~\cite{Wang2016}, which employs KPCA and 3D KPCA with an AdaBoost classifier, achieves high accuracy on the UCI dataset but remains limited to distinguishing falls from basic activities such as sitting, walking, and lying. It also relies on a small number of participants and exhibits suboptimal sensitivity despite the use of 3D KPCA. In contrast, our HAR framework captures six diverse and complex activities, demonstrating higher generalizability and interpretability. Similarly, the model proposed by Thakur \emph{et al.}~\cite{Thakur2021}, which combines PCA and t-SNE with an ensemble of RF, SVM, and LR classifiers, effectively visualizes data but struggles to capture nonlinear dependencies and incurs high computational cost. UMAP improves upon t-SNE in efficiency but still requires iterative neighbor graph construction and stochastic optimization, which restrict incremental learning and interpretability~\cite{Deng2024}. In comparison, CPE-IPCA efficiently captures nonlinear dependencies through polynomial expansion and supports incremental learning, while PPSA balances the influence of all principal components after projection, maintaining interpretability and low computational cost. These characteristics make our PPSA- and CPE-IPCA-based HAR framework superior in accuracy, scalability, and real-time feasibility compared to KPCA, t-SNE, and UMAP-based methods~\cite{Wang2016,Thakur2021,Deng2024}.
\section{Conclusion}
This study introduced a comprehensive Human Activity Recognition (HAR) framework using Ultra-Wideband (UWB) data consisting of 17,063 observations and 2,040 attributes. The framework effectively addressed challenges of high dimensionality and nonlinear dependencies through an integrated, interpretable, and computationally efficient pipeline combining data preprocessing, dimensionality reduction, association rule mining, and classification. Two novel dimensionality reduction methods were developed: Clustered Polynomial Expansion with Incremental PCA (CPE-IPCA) and the Post-PCA Standardization Approach for Nonlinear Data (PPSA-NLD).

CPE-IPCA clusters correlated features, captures nonlinear dependencies through polynomial expansion, and applies incremental PCA for scalable compression, producing a compact 50-dimensional representation. PPSA-NLD standardizes PCA-projected components for consistent scaling and improved interpretability, yielding an 80-dimensional feature space. Both reduced datasets were analyzed using Apriori and FP-Growth algorithms to extract interpretable, rule-based patterns and classified using Random Forest (RF) and the Vector Space Model (VSM).

Experimental results demonstrated exceptional performance: CPE-IPCA achieved 100\% accuracy with RF and F1-scores of 0.99–1.00 with VSM, while PPSA-NLD attained 99\% accuracy with RF and perfect precision, recall, and F1 with VSM. The proposed framework reliably classified six activities, Standff, Liedown, Noactivity, Sit, Stand, and Walk, demonstrating high accuracy, interpretability, and scalability for real-time HAR applications in smart environments, healthcare, and assistive technologies

\end{document}